\documentclass[11pt]{article}

\usepackage[T1]{fontenc}
\usepackage[utf8]{inputenc}
\usepackage[english]{babel}
\usepackage[margin=.92in]{geometry}
\usepackage{microtype}
\usepackage[numbers]{natbib}
\usepackage[colorlinks=true,linkcolor=blue,citecolor=blue,urlcolor=blue]{hyperref}
\usepackage{url}

\usepackage{amsmath,amssymb,amsthm,mathtools,mathrsfs}
\usepackage{array,booktabs,tabularx,makecell,longtable}
\usepackage{graphicx,subcaption,float}
\usepackage{xcolor}
\usepackage{enumitem}
\usepackage{algorithm}
\usepackage{algpseudocode}
\usepackage[toc,page]{appendix}
\usepackage{cancel}
\usepackage{placeins}

\theoremstyle{plain}
\newtheorem{thm}{Theorem}[section]
\newtheorem{cor}[thm]{Corollary}
\newtheorem{lem}[thm]{Lemma}

\theoremstyle{definition}
\newtheorem{defn}[thm]{Definition}
\newtheorem{assumption}[thm]{Assumption}

\theoremstyle{remark}
\newtheorem{rem}[thm]{Remark}

\newcommand{\R}{\mathbb{R}}

\newcommand{\PP}{\mathbb{P}}
\newcommand{\op}{\mathrm{op}}

\newcommand{\acc}{\operatorname{acc}}
\newcommand{\1}{\mathbf{1}}

\newcommand{\di}{\mathrm{d}}
\newcommand{\proofloc}[1]{\par\noindent\emph{Proof.} #1\par}

\newcommand{\maybeincludegraphics}[2][]{%
  \IfFileExists{#2}{\includegraphics[#1]{#2}}{%
  \fbox{\parbox[c][2.2in][c]{0.92\linewidth}{\centering Missing figure: \detokenize{#2}}}}}

\title{\textbf{Pruning Deep Neural Networks via the Marchenko--Pastur Distribution}}

\author{
    Leonid Berlyand \\
    {\small Department of Mathematics} \\
    {\small Pennsylvania State University} \\
    {\small University Park, PA 16802, USA}
    \and
    Theo Bourdais\\
    {\small Department of Computing and Mathematical Sciences} \\
    {\small California Institute of Technology} \\
    {\small Pasadena, CA 91125, USA}
    \and
    Houman Owhadi\\
    {\small Department of Computing and Mathematical Sciences} \\
    {\small California Institute of Technology} \\
    {\small Pasadena, CA 91125, USA}
    \and
    Yitzchak Shmalo\thanks{Corresponding author: Yitzchak Shmalo. Email: \texttt{yitzchak.shmalo@gmail.com}.} \\
    {\small Department of Mathematics} \\
    {\small Pennsylvania State University} \\
    {\small University Park, PA 16802, USA}
}
\date{}

\newcommand{\repoURL}{\url{https://github.com/yspennstate/RMT_based_pruning_in_deep_learning}}
\begin{document}
\maketitle

\begin{abstract}
We study a Marchenko--Pastur (MP) random-matrix approach to pruning deep neural networks with very small post-pruning fine-tuning budgets. The main practical contribution is accuracy retention under short calibration and fine-tuning schedules, rather than a long post-pruning reoptimization pipeline. The theory gives deterministic data-path certificates: if the removed component \(R\) has small propagated logit effect \(L_s\|R\psi_1(s)\|_\infty\), pruning decreases an elastic-net objective and preserves samples whose dense margin exceeds twice the perturbation. The zero-budget case gives perfect pruning; a prune--restore extension models weight restoration inside a fixed sparse-execution pattern; and an additive \(L_2\)-regularized model shows admissible random-like components vanish at the training limit, with persistent spikes stabilizing as the MP bulk collapses. Under iid-Gaussian sufficient conditions, the fitted MP edge \(\sigma_+\) gives a high-probability layerwise budget signal.

On ImageNet-1k, after only three distillation epochs, ViT-B/16 \(2{:}4{+}\)ToMe reaches \(83.41\%\) top-1 (\(-1.70\) pp from dense) at \(59.81\%\) sparse-execution MAC reduction, with \(1.388\times\) best-observed A40 native-\(2{:}4\) backend speedup for the same checkpoint and ToMe graph; a separate no-ToMe A100 endpoint gives \(2.705\times\). At structured sparsity, ViT-B/16 \(6{:}12\) reaches \(83.74\%\), ViT-L/16 \(8{:}16\) dense+permutation reaches \(85.33\%\) (\(-0.51\) pp), and ConvNeXtV2-Base \(12{:}16\) reaches \(86.35\%\) (\(-0.37\) pp). For CNNs, ResNet50 \(8{:}16\) dense+permutation reaches \(75.87\%\) (\(-0.26\) pp), and ResNet152d CAST-conv+permutation reaches \(81.33\%\) (\(-1.53\) pp) at \(\sim50\%\) MAC accounting with a \(1.62\times\) A40 im2col+\(2{:}4\) sparse-GEMM audit.
\end{abstract}


\noindent\textbf{Supplementary information.}
Full proofs, Gaussian/random-matrix specializations, and mathematical corollaries are provided in Online Resource 1. Full methodology, run recipes, checkpoint ledgers, timing-audit details, and the complete comparison table are provided in Online Resource 2. Mirror copies of the manuscript sources and PDFs are maintained at \repoURL.

\noindent\textbf{Norm convention.}
Throughout, \(\|x\|_\infty\) denotes the vector maximum norm, while \(\|A\|_\infty\) denotes the induced matrix norm
\[
\|A\|_\infty:=\max_i\sum_j |A_{ij}|.
\]

\section{Introduction}

DNN compression is motivated by overfitting, regularization, and deployment constraints. Random Matrix Theory (RMT) has been used to study trained spectra, implicit self-regularization, generalization diagnostics, Jacobians, and initialization \cite{martin2021implicit,mahoney2019traditional,meng2023impact,xiao2023heavy,thamm2022random,pastur2020random,pastur2023random,saada2023initialisation,martin2021predicting,martin2020heavy}. For fixed pruning hyperparameters, this paper uses MP spectral diagnostics to allocate masks without validation or test access at mask-construction time.

A main contribution is that the reported drops are obtained with little post-pruning fine-tuning: one epoch per unstructured pruning cycle and three distillation epochs for CAST/CAST-conv, rather than long retraining schedules. The empirical target is sparsity inside dense affine maps of Vision Transformers and related ImageNet models \cite{dosovitskiy2021image,vaswani2017attention}. Trained spectra are heterogeneous: the method treats MP-like layers as candidates for larger pruning budgets, while protecting non-MP or heavy-tailed layers. The main numerical results are deliberately front-loaded. On ImageNet-1k, Hybrid Magnitude--SER keeps ViT-B/16 at \(83.37\%\) top-1 at \(50\%\) unstructured sparsity, while the deployable CAST \(2{:}4{+}\)ToMe row reaches \(83.41\%\) after only three distillation epochs at \(59.81\%\) sparse-execution MAC reduction. The same checkpoint and ToMe graph gives a measured \(1.36\times\) fixed-batch A40 native-\(2{:}4\) speedup (\(1.388\times\) best batch-sweep value); a separate no-ToMe ViT-B/16 dense-to-\(2{:}4\) A100 endpoint gives \(2.705\times\). Wider structured projections improve the accuracy side: ViT-B/16 \(6{:}12\) reaches \(83.74\%\), ViT-L/16 \(8{:}16\) dense+permutation reaches \(85.33\%\) (\(-0.51\) pp), and ConvNeXtV2-Base \(12{:}16\) reaches \(86.35\%\) (\(-0.37\) pp). For CNNs, ResNet50 \(8{:}16\) dense+permutation reaches \(75.87\%\), only \(0.26\) pp below dense, and ResNet152d CAST-conv+permutation reaches \(81.33\%\), \(1.53\) pp below dense, with a \(1.62\times\) A40 im2col+\(2{:}4\) sparse-GEMM audit. The wider \(6{:}12\), \(8{:}16\), and \(12{:}16\) rows are accuracy/MAC-accounting rows, not native sparse Tensor Core throughput claims.

We also present three main theoretical results. First, the deterministic data-path certificate says that if the removed component \(R\) has small propagated logit effect \(L_s\|R\psi_1(s)\|_\infty\), then pruning decreases an elastic-net objective and preserves every training sample whose dense margin is larger than twice this perturbation (Lemma~\ref{lem:deterministic_ce_path_bound}, Corollary~\ref{cor:deterministic_margin_stability}, Theorem~\ref{thm:deterministic_mask_certificate}). In the zero-budget or ``perfect pruning'' case, the margin bound gives no accuracy loss on the training set. Second, the prune--restore certificate models structured \(k{:}n\) sparsity: restoring entries inside an already-paid sparse-execution group can improve the certificate while leaving the final sparse-execution pattern unchanged (Theorem~\ref{thm:deterministic_prune_restore_certificate}). Third, the additive \(L_2\)-regularized theory shows that, under one-sided stationarity and local-path convergence, admissible random-like components vanish at the training limit; the associated MP bulk collapses while persistent signal spikes stabilize (Theorem~\ref{thm:main_gaussian_stationarity_collapse}, Corollaries~\ref{cor:main_variance_scale_collapse}--\ref{cor:main_gaussian_BBP_persistent_spikes}, and the generalized Theorem~\ref{main_result_reducing_noise_general}). The MP edge enters as a sufficient random-matrix condition for data-path budgets and as the empirical layer-allocation signal used by SER/CAST. Tables~\ref{tab:vitb_current_methods}, \ref{tab:multi_arch_hybrid}, \ref{tab:param_to_flop_followup}, and~\ref{tab:deployability_audit_compact} report the main unstructured, structured, MAC-reduction, and deployability results; Table~\ref{tab:result_comparison} gives literature context. Prior pruning taxonomies and ViT/CNN compression citations are collected in Online Resource 2.

Section~\ref{sec:randomness} gives DNN and MP preliminaries. Section~\ref{sec:pruning_VITs} reports the numerical evidence. Sections~\ref{sec:abstract_additive_extensions} and~\ref{Main_theory_gen} state the additive and deterministic certificates. Full proofs and mathematical details are in Online Resource 1; full protocols, algorithms, provenance notes, and supplementary numerical material are in Online Resource 2.

\section{Randomness in Deep Neural Networks}
\label{sec:randomness}

\subsection{Introduction to Deep Neural Networks}
\label{introduction_deep_neural_networks}

In classification tasks, the goal is to assign each element of a set \(S\) to one of \(K\) classes. Let \(C(s)\in\{1,\dots,K\}\) denote the correct class of \(s\in S\). Given a labeled training set \(T\subset S\), we seek a classifier that generalizes from \(T\) to unseen data. We consider DNNs of the form
\[
\phi(\cdot,\alpha)=\rho\circ X(\cdot,\alpha),
\]
where \(\rho\) is the softmax map and \(X(\cdot,\alpha)\) is a composition of affine maps and nonlinearities:
\[
X(\cdot,\alpha)=\lambda\circ M_L(\cdot,\alpha)\circ\cdots\circ\lambda\circ M_1(\cdot,\alpha).
\]
Here:
\begin{itemize}[leftmargin=2em]
\item \(M_k(\cdot,\alpha)\) is an affine map \(\R^{N_{k-1}}\to\R^{N_k}\) with weight matrix \(W_k\in\R^{N_k\times N_{k-1}}\) and bias vector \(\beta_k\in\R^{N_k}\), so \(M_k(x)=W_kx+\beta_k\).
\item \(\lambda:\R^m\to\R^m\) is a nonlinear activation. In the simplified theory below we take \(\lambda\) to be either the coordinatewise absolute value or ReLU.
\item The softmax map \(\rho:\R^K\to\R^K\) is given by
\begin{equation}
\label{soft_max_new}
\rho(v)_i=\frac{e^{v_i}}{\sum_{j=1}^{K} e^{v_j}},\qquad v\in\R^K.
\end{equation}
\end{itemize}
The standard cross-entropy loss is
\begin{equation}
\label{CE}
L_{\mathrm{CE}}(\alpha)=-\frac{1}{|T|}\sum_{s\in T}\log\big(\phi_{C(s)}(s,\alpha)\big).
\end{equation}

\subsection{The MP distribution in machine learning contexts}
\label{sec:MP_distribution}

The Marchenko--Pastur distribution is a basic object in RMT \cite{marchenko1967distribution}; high-dimensional random-matrix methods more broadly have applications in signal processing, statistics, wireless communication, and machine learning \cite{vershynin2018high,ge2021large,serdobolskii2000multivariate,couillet2011random}. We first define the relevant empirical spectral distributions.

\begin{defn}[Eigenvalue and singular-value empirical spectral distributions]
\label{ESD_Definition}
Let \(G\in\R^{N\times M}\), and let \(s_1(G),\dots,s_{\min\{N,M\}}(G)\) denote its singular values, counted with multiplicity and including possible zeros. The singular-value empirical spectral distribution (ESD) of \(G\) is
\[
\nu_G:=\frac{1}{\min\{N,M\}}\sum_{i=1}^{\min\{N,M\}}\delta_{s_i(G)}.
\]
If \(A\in\R^{M\times M}\) is symmetric positive semidefinite with eigenvalues \(\lambda_1(A),\dots,\lambda_M(A)\), its eigenvalue ESD is
\[
\mu_A:=\frac{1}{M}\sum_{i=1}^M \delta_{\lambda_i(A)}.
\]
\end{defn}

\begin{thm}[Marchenko--Pastur law]
\label{RMT_MP_theorem}
Let \(W_N\) be an \(N\times M_N\) random matrix with i.i.d.\ entries of mean \(0\), variance \(\sigma^2\), and finite fourth moment. Define
\[
X_N:=\frac{1}{N}W_N^\top W_N,
\qquad c_N:=\frac{M_N}{N}\to c\in(0,\infty).
\]
Then the eigenvalue ESD \(\mu_{X_N}\) converges almost surely to the Marchenko--Pastur law
\begin{equation}
\label{MP_distribution}
\mu_{\mathrm{MP}}^{c,\sigma^2}
=
\left(1-\frac{1}{c}\right)_+\delta_0
+
\frac{\sqrt{(\lambda_+-x)(x-\lambda_-)}}{2\pi c\sigma^2 x}\,
\1_{[\lambda_-,\lambda_+]}(x)\,\di x,
\end{equation}
where
\begin{equation}
\label{lambda_parameters}
\lambda_\pm=\sigma^2(1\pm \sqrt c)^2.
\end{equation}
\end{thm}
\proofloc{This is a classical cited random-matrix input rather than a new theorem of the paper; see Online Resource 1 for source notes.}

\begin{rem}
If \(M_N\le N\) for all \(N\), then \(c\in(0,1]\) and the atom at zero in \eqref{MP_distribution} vanishes. When discussing the singular-value scale of \(W_N\), we write
\[
\sigma_+ := \sqrt{N\lambda_+}
\]
for the corresponding MP upper edge.
\end{rem}

\subsection{Reduction of randomness in DNN weights via MP diagnostics}
\label{overallpicture}

At common initialization scale \(W_{ij}\) has variance \(g/N\), so the remark above gives \(\sigma_+=\sqrt g\,(1+\sqrt c)\) in the rectangular aspect-ratio limit \(M/N\to c\). Singular values and the nonzero eigenvalues of \(W^\top W\) are then \(O(1)\), while the eigenvalues of \(X_\ell=W_\ell^\top W_\ell/N\) are \(O(1/N)\); learned spectra later deviate from the initial random law \cite{martin2021implicit,staats2022boundary}. Training introduces structure, motivating the following signal-plus-randomness suppositions.

\paragraph{Supposition 1.}
After \(t\) training steps, the layer-\(\ell\) weight matrix can be decomposed as
\[
W_\ell(t)=R_\ell(t)+S_\ell(t),
\]
where \(R_\ell(t)\) is an independent random perturbation and \(S_\ell(t)\) is a structured signal component.

\paragraph{Supposition 2.}
For the spectral spike interpretation, \(S_\ell(t)\) is low rank or approximately low rank relative to the width.

\begin{figure}[t]
\centering
\begin{subfigure}[t]{0.48\textwidth}
\centering
\maybeincludegraphics[width=\linewidth]{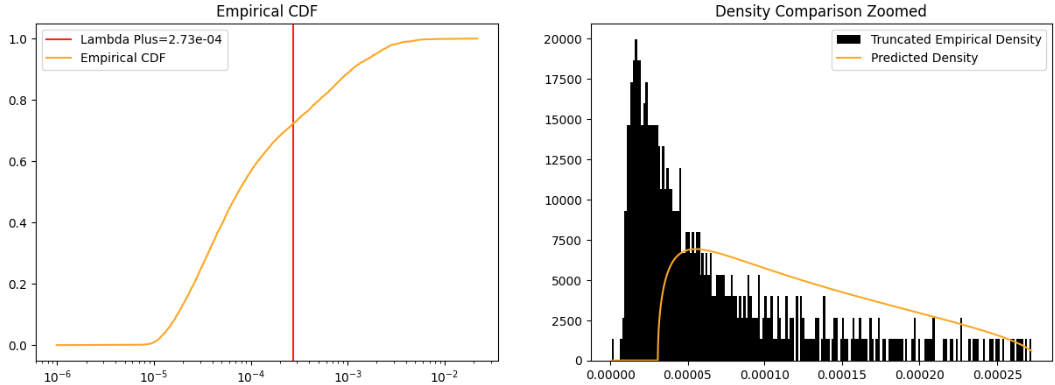}
\caption{ViT-B/16 layer example with MP-fit error \(0.74\) and bulk fraction \(73\%\).}
\label{fig:vit_mp_poor_fit}
\end{subfigure}\hfill
\begin{subfigure}[t]{0.48\textwidth}
\centering
\maybeincludegraphics[width=\linewidth]{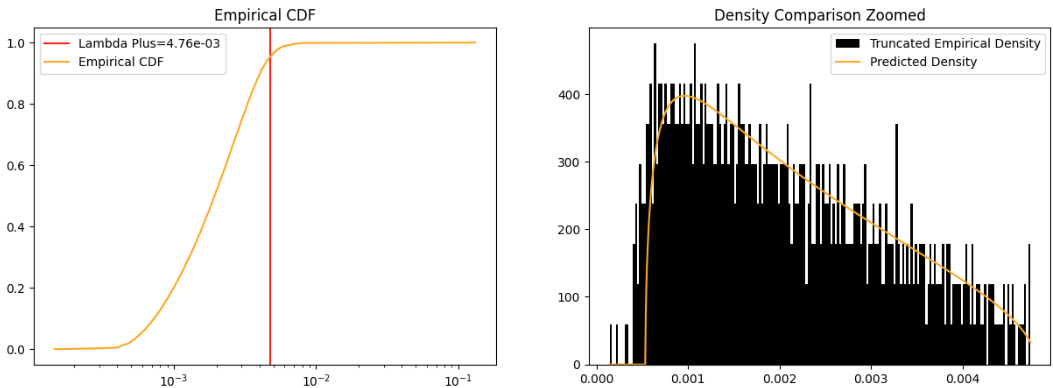}
\caption{ViT-B/16 layer example with MP-fit error \(0.01\) and bulk fraction \(99.73\%\).}
\label{fig:vit_mp_strong_fit}
\end{subfigure}
\caption{Layerwise MP diagnostics for two projection matrices from the same trained ViT-B/16.}
\label{fig:vit_mp_diagnostics}
\end{figure}

These suppositions are modeling devices for the MP-budget argument, not assertions that trained ViTs are globally iid noise plus low-rank signal. Figure~\ref{fig:vit_mp_diagnostics} gives two representative layer diagnostics; the assumptions behind using MP fit for budgets are stated in Sections~\ref{sec:abstract_additive_extensions} and~\ref{Main_theory_gen}.

\section{Numerical evidence: RMT-guided pruning of ViTs}
\label{sec:pruning_VITs}

This section reports empirical pruning evidence and gives the minimum algorithmic context needed to read the tables.  Full run recipes, scripts, checkpoint ledgers, calibration details, and migrated numerical appendices are in Online Resource 2.  All ImageNet-1k top-1 values in this section are measured after the stated pruning and fine-tuning recipe on the corresponding dense checkpoint; validation labels are not used to construct masks.  The dense baseline in a row is therefore part of the row's protocol, and comparisons to external work should be read through the reported drop, compression axis, training budget, and deployment note rather than through raw top-1 alone.

The simplest baseline is magnitude pruning.  For a target sparsity \(s\), it removes the entries with the smallest absolute weights in the prunable tensors, either globally or within the layer set specified by the experiment.  Magnitude pruning is useful because it is deterministic, cheap, and hard to dismiss: if an RMT-guided method cannot beat it under the same checkpoint, layer set, and fine-tuning budget, then the spectral signal has not added evidence.  It is also incomplete for this setting.  At high sparsity it can spend too much budget in layers whose small entries still matter along the data path, and it has no mechanism for allocating different budgets from the layer spectra.

The ``classical RMT'' baseline is different.  It fits a Marchenko--Pastur bulk to a weight matrix or to a reshaped convolutional tensor, estimates an upper bulk edge \(\sigma_+\), and treats the sub-edge component as the random-like part suggested by the fitted spectrum.  In its direct pruning form, the method removes or sparsifies this component by singular-vector reconstruction rather than by weight magnitude alone.  This is an important diagnostic baseline because it asks whether the MP edge by itself identifies removable directions.  Table~\ref{tab:vitb_current_methods} shows that it helps relative to magnitude at some higher sparsities, but it is not the final method: a literal sub-edge reconstruction is too coarse for modern ViT weights, and the paper does not claim that every sub-edge direction is noise.

The practical unstructured method is SER, short for sparsify--estimate--restore.  SER starts from an over-pruned candidate, evaluates the lost mass through the same MP-guided budget view, and restores entries that are most useful for the target sparse model.  The point is not to keep the smallest MP-bulk entries automatically; it is to use the MP fit as a layerwise budget signal, then let the restoration step decide which entries should return.  Hybrid Magnitude--SER uses plain magnitude pruning at low sparsity, where it is strong and stable, then switches to SER once the pruning level is high enough that layer allocation and restoration matter.  In the canonical ViT-B/16 row this gives \(83.37\%\) top-1 at \(50\%\) unstructured sparsity, compared with \(81.67\%\) for magnitude and \(82.57\%\) for the direct RMT baseline under the same dense checkpoint and short post-pruning fine-tuning convention.

CAST is the structured-sparsity version of this idea.  The target is a fixed \(k{:}n\) pattern: in every group of \(n\) weights, \(k\) entries are kept and the rest are zeroed.  A \(2{:}4\) pattern is the deployable NVIDIA sparse-tensor-core case; wider \(6{:}12\), \(8{:}16\), and \(12{:}16\) patterns are accuracy/MAC-accounting probes unless a native backend for that exact pattern is explicitly audited.  CAST scores candidate group patterns using a certificate-inspired objective: preserve entries that matter for the propagated data-path perturbation while respecting the final group sparsity.  The ``free restoration'' terminology means that, inside a group whose final sparse-execution cost is already fixed, restoring a kept entry can improve the certificate without changing the number of executed sparse weights.

The CAST rows should therefore be split into two categories.  Rows marked \(2{:}4\) with native sparse-kernel audits are deployment evidence: the same final pattern can be represented by a backend that has a real sparse execution path.  The ViT-B/16 \(2{:}4{+}\)ToMe checkpoint, for example, reaches \(83.41\%\) top-1 at \(59.81\%\) sparse-execution MAC reduction and has a measured A40 native-\(2{:}4\) speedup for the same checkpoint and ToMe graph.  The ResNet rows use a different audit: convolution is lowered to im2col and then timed as a \(2{:}4\) sparse GEMM endpoint, so those numbers are evidence that the weight pattern can accelerate the lowered matrix multiply, not proof of an end-to-end cuDNN Conv2d speedup.  The wider \(k{:}n\) rows answer a separate question: how much accuracy is available if the structured group is less constrained than \(2{:}4\), or if a future backend exposes a wider pattern.

Token merging is another axis and is held separate in the interpretation.  ToMe reduces the number of tokens processed by the ViT graph; \(2{:}4\) pruning reduces the executed weights inside eligible linear maps.  The \(2{:}4{+}\)ToMe rows report their combined sparse-execution MAC accounting because that is the graph being evaluated, but the backend speedup claims in Table~\ref{tab:deployability_audit_compact} specify whether ToMe is held fixed.  This distinction is why the different no-ToMe A100 dense-to-\(2{:}4\) endpoint is reported separately in Online Resource 2 rather than pooled with the A40 ToMe-held-fixed audits.

The result blocks have the following roles.  Table~\ref{tab:vitb_current_methods} is the matched ViT-B/16 ablation comparing magnitude, direct RMT, SER, and Hybrid Magnitude--SER.  Table~\ref{tab:multi_arch_hybrid} tests whether the Hybrid pattern survives across ViT, Swin, ConvNeXt, ResNet, and Hiera checkpoints at the same sparsity grid.  Table~\ref{tab:param_to_flop_followup} reports the structured \(k{:}n\) rows used for the main MAC-reduction claims.  Table~\ref{tab:deployability_audit_compact} is the speed audit.  The comparison table gives the closest available published contexts without treating them as a controlled leaderboard.  The full theory-to-numerics map is in Online Resource 2; it matches each CAST/SER implementation step to the lemma, theorem, corollary, or empirical proxy it is meant to instantiate, and it states explicitly where the link is motivational rather than a verified certificate.  The theoretical sections then explain why the algorithms are phrased as certificate-inspired pruning: the deterministic lemmas control the data-path effect of a removed component, while MP provides a sufficient random-matrix condition and a practical layerwise signal, not a complete certified transformer bound.

\subsection{ViT-B/16 results}
\label{subsec:vitb_current_results}

\begin{table}[!htbp]
\centering
\caption{ViT-B/16 pruning results on ImageNet-1k validation under this paper's matched checkpoint, prunable-layer set, sparsity grid, evaluation pipeline, and short fine-tuning budget. Entries are top-1 accuracy (\%) at sparsity \(s\); the dense baseline is \(85.11\%\). Bold marks the highest entry in each sparsity column.}
\label{tab:vitb_current_methods}
\scriptsize
\setlength{\tabcolsep}{2.5pt}
\begin{tabular}{lcccccccccccccc}
\toprule
Method & \(0.05\) & \(0.10\) & \(0.15\) & \(0.20\) & \(0.25\) & \(0.30\) & \(0.35\) & \(0.40\) & \(0.45\) & \(0.50\) & \(0.55\) & \(0.60\) & \(0.65\) & \(0.70\) \\
\midrule
Classical magnitude   & 85.21 & 85.16 & \textbf{85.08} & 84.87 & 84.65 & 84.46 & 84.08 & 83.53 & 82.86 & 81.67 & 80.17 & 77.98 & 74.30 & 67.44 \\
Classical RMT         & 85.21 & 85.14 & 85.01 & \textbf{84.97} & 84.77 & 84.55 & 84.31 & 83.98 & 83.31 & 82.57 & 81.31 & 79.13 & 76.46 & 71.53 \\
SER                   & 84.94 & 84.98 & 84.82 & 84.76 & 84.66 & 84.55 & 84.41 & 84.24 & 83.73 & 83.27 & 82.65 & 81.33 & \textbf{79.95} & 77.94 \\
Hybrid Magnitude--SER & \textbf{85.23} & \textbf{85.17} & 85.06 & 84.89 & \textbf{84.81} & \textbf{84.64} & \textbf{84.55} & \textbf{84.28} & \textbf{83.80} & \textbf{83.37} & \textbf{82.76} & \textbf{81.39} & \textbf{79.95} & \textbf{78.01} \\
\bottomrule
\end{tabular}
\end{table}

Layerwise interpretation and mask-only results are in Online Resource 2.

\subsection{Broader architecture sweep}
\label{subsec:architecture_validation}

Table~\ref{tab:multi_arch_hybrid} gives the Hybrid Magnitude--SER sweep; Online Resource 2 gives architecture provenance, adaptive-variant details, slope/certificate-audit analyses, aggregate statistics, and checkpoint ledgers.

\begin{table}[!htbp]
\centering
\caption{Hybrid Magnitude--SER results across ImageNet-1k checkpoints. Entries are post-FT top-1 accuracy (\%) at sparsity \(s\). ``Base.'' is the dense top-1 measured in this paper's evaluation pipeline. Bold marks selected notable \(s=0.50\) entries: small drops from dense, high-accuracy large-model rows, or strong CNN coverage. Online Resource 2 retains the protocol markers and detailed notes for the adaptive rows.}
\label{tab:multi_arch_hybrid}
\tiny
\setlength{\tabcolsep}{1.0pt}
\begin{tabular}{lccccccccccccccc}
\toprule
Architecture (timm checkpoint) & Base. & \(0.05\) & \(0.10\) & \(0.15\) & \(0.20\) & \(0.25\) & \(0.30\) & \(0.35\) & \(0.40\) & \(0.45\) & \(0.50\) & \(0.55\) & \(0.60\) & \(0.65\) & \(0.70\) \\
\midrule
ViT-B/16 \texttt{augreg2\_in21k\_ft\_in1k} & 85.11 & 85.23 & 85.17 & 85.06 & 84.89 & 84.81 & 84.64 & 84.55 & 84.28 & 83.80 & \textbf{83.37} & 82.76 & 81.39 & 79.95 & 78.01 \\
ViT-B/16/384 \texttt{augreg\_in21k\_ft}    & 86.01 & 86.05 & 86.09 & 86.07 & 85.99 & 86.00 & 85.90 & 85.81 & 85.78 & 85.48 & \textbf{85.15} & 84.64 & 83.35 & 82.50 & 81.33 \\
ViT-Large/16 \texttt{augreg\_in21k\_ft}    & 85.84 & 85.80 & 85.84 & 85.88 & 85.75 & 84.98 & 85.32 & 85.64 & 85.04 & 85.08 & \textbf{84.52} & 84.34 & 83.81 & 83.02 & 82.13 \\
DeiT-Tiny \texttt{patch16\_224.fb\_in1k}   & 72.21 & 72.41 & 72.38 & 72.18 & 71.86 & 71.99 & 71.75 & 71.35 & 70.66 & 68.97 & 67.74 & 65.91 & 61.17 & 55.58 & 52.55 \\
DeiT-Small \texttt{patch16\_224.fb\_in1k}  & 79.85 & 79.82 & 79.78 & 79.62 & 79.37 & 79.31 & 79.21 & 79.00 & 78.61 & 78.06 & 77.22 & 76.25 & 74.14 & 72.05 & 68.55 \\
DeiT-Base \texttt{patch16\_224.fb\_in1k}   & 81.80 & 81.76 & 81.70 & 81.58 & 81.46 & 81.42 & 81.28 & 81.17 & 80.99 & 80.62 & 80.12 & 79.49 & 78.16 & 76.72 & 74.90 \\
Swin-Tiny \texttt{patch4\_window7\_224}    & 81.20 & 81.32 & 81.28 & 81.25 & 81.05 & 81.09 & 80.91 & 80.78 & 80.37 & 80.14 & 79.80 & 78.98 & 78.06 & 76.64 & 74.47 \\
ConvNeXt-Base \texttt{in22k\_ft\_in1k}     & 85.84 & 85.72 & 85.58 & 85.51 & 85.26 & 85.35 & 85.39 & 85.26 & 85.04 & 84.94 & \textbf{84.80} & 84.09 & 83.70 & 83.11 & 81.21 \\
ResNet50d \texttt{ra2\_in1k}               & 80.55 & 79.90 & 80.01 & 80.02 & 80.12 & 80.09 & 80.12 & 80.06 & 79.90 & 79.48 & 79.16 & 78.93 & 77.93 & 77.25 & 76.32 \\
ResNet101d \texttt{ra2\_in1k}              & 82.26 & 82.06 & 82.05 & 82.15 & 82.15 & 82.06 & 82.07 & 81.93 & 81.97 & 81.61 & \textbf{81.56} & 81.47 & 80.45 & 80.15 & 79.82 \\
Hiera-Base+ \texttt{mae\_in1k\_ft\_in1k}   & 84.40 & 85.04 & 84.94 & 84.76 & 84.39 & 77.95 & 83.12 & 82.15 & 82.37 & 82.29 & 82.25 & 82.00 & 80.03 & 80.55 & 78.96 \\
ResNet18 \texttt{tv\_in1k}    & 69.76 & 69.73 & 69.52 & 69.18 & 68.94 & 69.02 & 69.28 & 69.50 & 69.50 & 69.39 & \textbf{69.12} & 69.00 & 68.31 & 67.82 & 67.23 \\
ResNet34 \texttt{tv\_in1k}    & 73.28 & 73.00 & 72.67 & 72.39 & 72.62 & 73.06 & 73.09 & 73.16 & 73.16 & 73.00 & \textbf{72.88} & 72.74 & 72.12 & 71.59 & 71.44 \\
ResNet50 \texttt{tv\_in1k}    & 76.13 & 75.85 & 75.33 & 75.72 & 76.07 & 76.17 & 76.13 & 76.13 & 76.14 & 75.75 & \textbf{75.76} & 75.70 & 74.83 & 74.62 & 74.15 \\
DeiT-Base \texttt{patch16\_224.fb\_in1k} & 81.97 & 81.79 & 81.69 & 81.54 & 81.44 & 81.25 & 81.35 & 81.18 & 80.95 & 80.68 & 80.12 & 79.48 & 78.24 & 76.70 & 74.78 \\
Swin-Tiny \texttt{patch4\_window7\_224} & 81.39 & 81.32 & 81.29 & 81.26 & 81.05 & 80.77 & 80.47 & 80.58 & 80.43 & 80.11 & 79.80 & 78.94 & 78.03 & 76.53 & 74.31 \\
ConvNeXtV2-Base \texttt{fcmae\_ft\_in22k\_in1k} & 86.72 & 86.67 & 86.58 & 86.43 & 86.27 & 85.93 & 85.97 & 85.96 & 85.71 & 85.58 & \textbf{85.33} & 84.81 & 84.61 & 83.83 & 81.40 \\
\bottomrule
\end{tabular}
\end{table}

\begin{figure}[H]
\centering
\includegraphics[width=\textwidth]{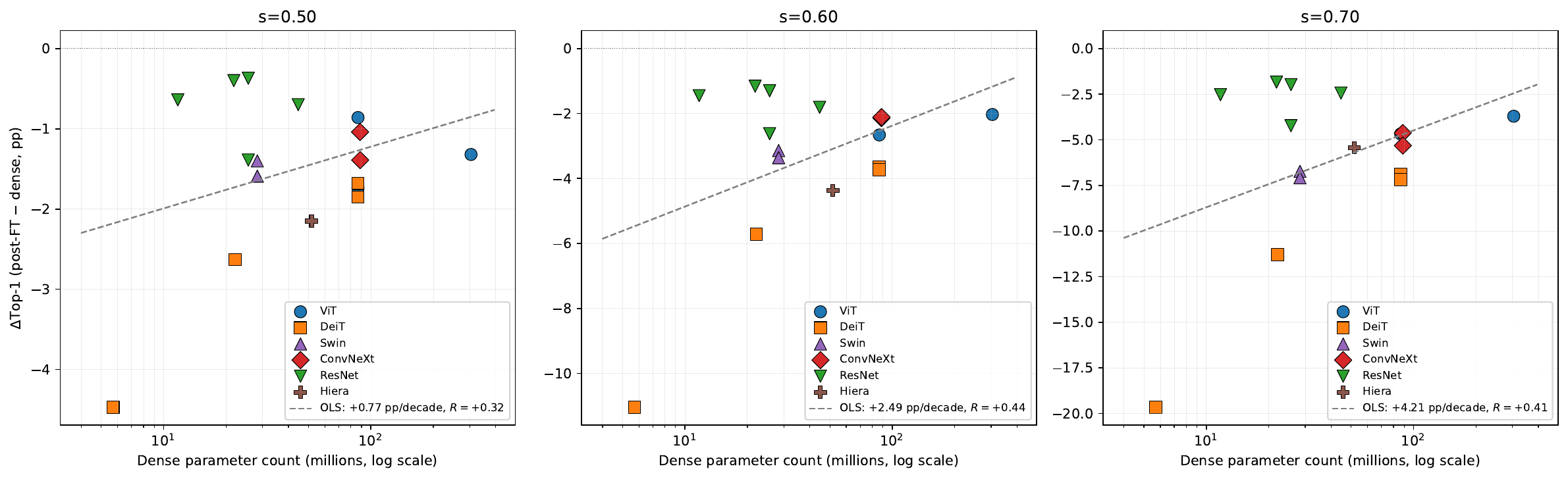}
\caption{\(\Delta\)Top-1 relative to each dense baseline after Hybrid Magnitude--SER pruning vs.\ dense parameter count. Points are Table~\ref{tab:multi_arch_hybrid} entries; marker style encodes architecture family. Dashed lines are OLS fits at the plotted sparsity levels.}
\label{fig:param_vs_drop}
\end{figure}

\begin{figure}[H]
\centering
\includegraphics[width=0.82\textwidth]{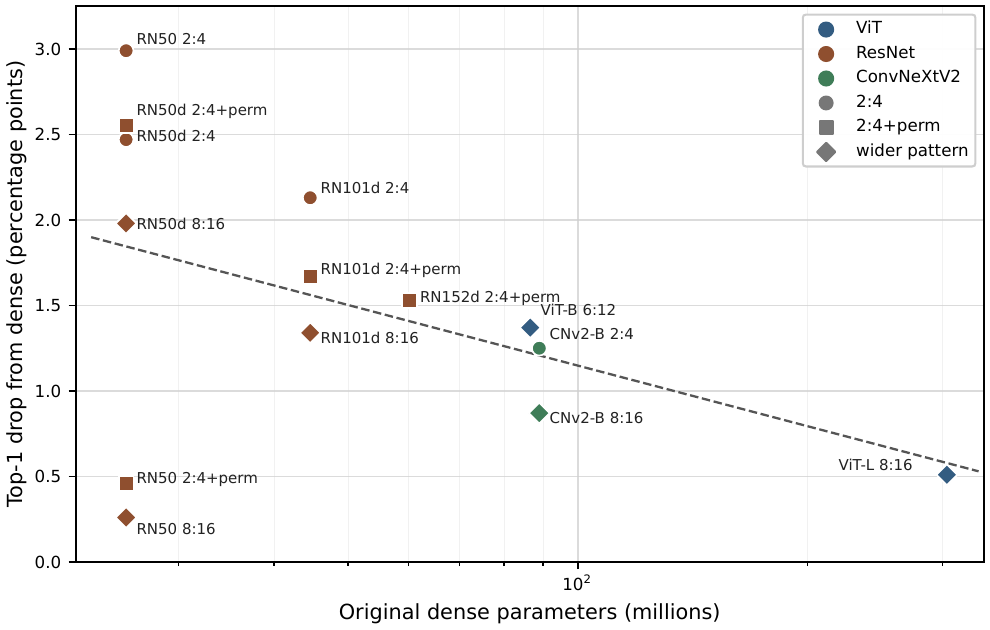}
\caption{Top-1 drop from dense vs.\ original dense parameter count for rows of Table~\ref{tab:param_to_flop_followup} at 48.5--50\% dense-equivalent sparse-execution MAC reduction. The y-axis is dense top-1 minus post-FT top-1, so lower is better. The dashed line is an OLS fit in \(\log_{10}\) parameters.}
\label{fig:flop_top1_vs_params}
\end{figure}

Figures~\ref{fig:param_vs_drop} and~\ref{fig:flop_top1_vs_params} show the same size trend in two views: at matched nominal sparsity or MAC reduction, larger dense checkpoints tend to pay a smaller top-1 penalty.  This is theoretically plausible for RMT-guided pruning, because the MP edge is an asymptotic spectral object that is better resolved in larger matrices, and because wider overparameterized layers can contain a larger random-like bulk reservoir whose removal has small data-path effect under the certificate model.  The figures are descriptive finite-model evidence for this scaling intuition, not a theorem that size alone guarantees pruneability.

\subsection{Cert-aware structured sparsity: the FLOP / speedup table}
\label{subsec:flop_table}

Table~\ref{tab:param_to_flop_followup} summarizes structured sparsity; The CAST algorithm and source/checkpoint recipes are in Online Resource 2.

\begin{table}[H]
\centering
\caption{CAST \(k{:}n\) structured-projection results. ``MAC red.'' is dense-equivalent MAC-accounting reduction; for rows without a deploy-speed entry this is not an audited sparse-backend wall-clock claim. ``Th. spd.'' is \(1/(1-\mathrm{MAC\ red.})\), and \(\Delta\) is top-1 minus the corresponding dense baseline. ``Deploy spd.'' lists the selected best-observed measured sparse-backend speedup when available; Table~\ref{tab:deployability_audit_compact} reports endpoint, batch, and audit details. Rows labeled ``(ours)'' are this paper's measurements. Bold marks notable accuracy or measured-speed entries, not row ownership. A dash means no exact sparse-backend wall-clock speedup is claimed.}
\label{tab:param_to_flop_followup}
\scriptsize
\setlength{\tabcolsep}{1.1pt}
\renewcommand{\arraystretch}{1.05}
\begin{tabularx}{\textwidth}{l c X c c c c c c}
\toprule
Architecture & Source \(s\) & Method & Dense MACs & MAC red. & Th. spd. & Deploy spd. & Top-1 (\%) & \(\Delta\) (pp) \\
\midrule
\multicolumn{9}{l}{\textit{ViT family}} \\
ViT-B/16   & 0.35 & Magnitude 2:4 + ToMe (ours) & 17.56G & 59.81\% & \(2.49\times\) & \(1.388\times\)\textsuperscript{*} & 82.92 & \(-2.19\) \\
ViT-B/16   & 0.35 & CAST 2:4 + ToMe (ours)        & 17.56G & 59.81\% & \(2.49\times\) & \(\mathbf{1.388\times}\) & \(\mathbf{83.41}\) & \(-1.70\) \\
ViT-B/16   & 0.35 & CAST 6:12 SER+\(\alpha\)=0.5 (ours, no ToMe) & 17.56G & 50\% & \(2.00\times\) & -- & \(\mathbf{83.74}\)\textsuperscript{\S} & \(\mathbf{-1.37}\) \\
ViT-L/16   & 0.35 & CAST 2:4 + ToMe (ours)                 & 61.55G\textsuperscript{\dag} & \(\sim\)60\%\textsuperscript{\dag} & \(2.50\times\) & \(\mathbf{1.394\times}\) & \(\mathbf{84.37}\) & \(-1.47\) \\
ViT-L/16   & dense & CAST 8:16 dense+perm (ours, no ToMe) & 61.55G & 50\% & \(2.00\times\) & -- & \(\mathbf{85.33}\)\textsuperscript{\S} & \(\mathbf{-0.51}\) \\
DeiT-B     & 0.35 & CAST 2:4 + ToMe                 & 17.56G & 59.81\% & \(2.49\times\) & \(1.384\times\) & 80.48 & \(-1.32\) \\
DeiT-S     & 0.35 & CAST 2:4 + ToMe                 & 4.61G  & 59.81\% & \(2.49\times\) & \(1.376\times\) & 76.96 & \(-2.89\) \\
DeiT-T     & 0.35 & CAST 2:4 + ToMe                 & 1.26G  & 59.81\% & \(2.49\times\) & \(1.330\times\) & 65.93 & \(-6.28\) \\
\midrule
\multicolumn{9}{l}{\textit{ResNet family}} \\
ResNet50   & 0.35 & CAST-conv          & 4.09G  & 48.5\%   & \(1.94\times\) & -- & 73.14 & \(-2.99\) \\
ResNet50   & 0.35 & CAST-conv+perm (ours) & 4.09G  & 48.5\%   & \(1.94\times\) & \(\mathbf{1.700\times}\)\textsuperscript{\ddag} & \(\mathbf{75.67}\)\textsuperscript{\ddag} & \(\mathbf{-0.46}\) \\
ResNet50   & dense & CAST 8:16 dense+perm (ours) & 4.09G  & 50\%     & \(2.00\times\) & -- & \(\mathbf{75.87}\)\textsuperscript{\S} & \(\mathbf{-0.26}\) \\
ResNet50d  & 0.35 & CAST-conv          & 4.33G  & 49.85\%  & \(1.99\times\) & -- & 78.08 & \(-2.47\) \\
ResNet50d  & 0.35 & CAST-conv+perm (ours) & 4.33G  & 49.85\%  & \(1.99\times\) & \(1.496\times\)\textsuperscript{\ddag} & 78.00\textsuperscript{\ddag} & \(-2.55\) \\
ResNet50d  & dense & CAST 8:16 dense+perm (ours) & 4.33G  & 50\%     & \(2.00\times\) & -- & \(\mathbf{78.57}\)\textsuperscript{\S} & \(-1.98\) \\
ResNet101d & 0.35 & CAST-conv          & 8.0G   & \(\sim\)50\% & \(2.00\times\) & -- & 80.13 & \(-2.13\) \\
ResNet101d & 0.35 & CAST-conv+perm (ours) & 8.0G   & \(\sim\)50\% & \(2.00\times\) & \(\mathbf{1.568\times}\)\textsuperscript{\ddag} & \(\mathbf{80.59}\)\textsuperscript{\ddag} & \(-1.67\) \\
ResNet101d & dense & CAST 8:16 dense+perm (ours) & 8.0G   & 50\%     & \(2.00\times\) & -- & \(\mathbf{80.92}\)\textsuperscript{\S} & \(-1.34\) \\
ResNet152d & 0.35 & CAST-conv+perm (ours) & 11.8G  & \(\sim\)50\% & \(2.00\times\) & \(\mathbf{1.617\times}\)\textsuperscript{\ddag} & \(\mathbf{81.33}\)\textsuperscript{\ddag} & \(-1.53\) \\
\midrule
\multicolumn{9}{l}{\textit{ConvNeXt family}} \\
ConvNeXtV2-Base & 0.35 & CAST 2:4 cert + free-restore (ours) & 15.4G & 50\% & \(2.00\times\) & \(1.295\times\) & \(\mathbf{85.47}\) & \(-1.25\) \\
ConvNeXtV2-Base & dense & CAST 12:16 dense+perm (ours, 25\% sparse) & 15.4G & 25\% & \(1.33\times\) & -- & \(\mathbf{86.35}\)\textsuperscript{\S} & \(\mathbf{-0.37}\) \\
ConvNeXtV2-Base & dense & CAST 8:16 dense+perm (ours, 50\% sparse) & 15.4G & 50\% & \(2.00\times\) & -- & \(\mathbf{85.85}\)\textsuperscript{\S} & \(\mathbf{-0.87}\) \\
\bottomrule
\end{tabularx}\\[2pt]
{\footnotesize \textsuperscript{\dag}Analytical ViT-L/16 ToMe+2:4 estimate. \textsuperscript{\ddag}Exact im2col+\(2{:}4\) sparse-GEMM audit, not faster cuDNN Conv2d. \textsuperscript{\S}Accuracy/MAC-accounting probe unless a measured endpoint is listed; dense-source wider-pattern rows are not incremental comparisons to the \(s=0.35\) source rows. \textsuperscript{*}Magnitude row shares the ViT-B 2:4+ToMe deployment path; see Online Resource 2.}
\end{table}

\begin{table}[H]
\centering
\caption{Deployability audit for the main measured checkpoints, reporting best-observed A40 batch-sweep values. Accuracy is fixed by the saved post-FT checkpoint; no weights are rewritten, and only the in-memory execution endpoint changes. Linear rows use the PyTorch/NVIDIA native \(2{:}4\) backend on A40. ResNet rows use exact im2col+\(2{:}4\) sparse GEMM relative to dense im2col; these rows do not claim a faster cuDNN Conv2d endpoint. Bold marks notable measured speedups, including the strongest endpoints within the Linear and ResNet audit groups.}
\label{tab:deployability_audit_compact}
\scriptsize
\setlength{\tabcolsep}{2.1pt}
\renewcommand{\arraystretch}{1.05}
\begin{tabularx}{\textwidth}{l l c r r c X}
\toprule
Checkpoint & Audited endpoint & Batch & Dense im/s & Sparse im/s & Speedup & Status \\
\midrule
ViT-B/16 + ToMe & native Linear 2:4 & 64 & 1191.8 & 1654.5 & \(\mathbf{1.388\times}\) & native sparse Tensor Core path \\
ViT-L/16 + ToMe & native Linear 2:4 & 64 & 620.8 & 865.6 & \(\mathbf{1.394\times}\) & native sparse Tensor Core path \\
DeiT-B + ToMe & native Linear 2:4 & 64 & 1189.6 & 1646.7 & \(1.384\times\) & native sparse Tensor Core path \\
DeiT-S + ToMe & native Linear 2:4 & 128 & 2708.5 & 3725.8 & \(1.376\times\) & native sparse Tensor Core path \\
DeiT-T + ToMe & native Linear 2:4 & 128 & 5158.4 & 6859.4 & \(1.330\times\) & native sparse Tensor Core path \\
ConvNeXtV2-B & native Linear 2:4 & 64 & 499.3 & 646.5 & \(1.295\times\) & pointwise Linear layers accelerated; dense convs unchanged \\
ResNet50 & im2col sparse GEMM & 128 & 469.5 & 797.9 & \(\mathbf{1.700\times}\) & im2col \(2{:}4\) sparse GEMM \\
ResNet50d & im2col sparse GEMM & 128 & 445.2 & 666.2 & \(\mathbf{1.496\times}\) & im2col \(2{:}4\) sparse GEMM \\
ResNet101d & im2col sparse GEMM & 128 & 228.2 & 357.8 & \(\mathbf{1.568\times}\) & im2col \(2{:}4\) sparse GEMM \\
ResNet152d & im2col sparse GEMM & 128 & 159.9 & 258.7 & \(\mathbf{1.617\times}\) & im2col \(2{:}4\) sparse GEMM \\
\bottomrule
\end{tabularx}
\end{table}

\paragraph{Deployability interpretation.}
Endpoint coverage and the separate A100 no-ToMe ViT-B benchmark are documented in Online Resource 2.

Several CAST rows start from denser, higher-accuracy checkpoints than the cited ResNet references. In that regime the remaining weights are plausibly carrying more useful signal, so the relevant comparison is the reported drop together with the training budget and measured endpoint, not raw top-1 alone. The wider \(6{:}12\), \(8{:}16\), and \(12{:}16\) accuracy/MAC probes remain in Table~\ref{tab:param_to_flop_followup}; Table~\ref{tab:result_comparison} carries the closest published unstructured and structured-sparsity context; the complete comparison table is provided in Online Resource 2.

\subsection{Comparison with published pruning and structured-sparsity baselines}
\label{subsec:result_comparison}

Table~\ref{tab:result_comparison} is literature context, not matched baseline evidence. Its main purpose is to show that the small accuracy drops reported here are obtained with very limited fine-tuning compared with many published pruning pipelines; matched evidence is Table~\ref{tab:vitb_current_methods}, Table~\ref{tab:multi_arch_hybrid}, and the caveats in Online Resource 2. Reference rows in this table and in the extended Online Resource 2 comparison table were checked against the corresponding source-paper tables; protocol-mismatched rows (different baseline checkpoint, different sparsity axis, different fine-tuning budget, or different deployability claim) are included only as context, not as apples-to-apples baselines. We separate unstructured or fine-grained weight sparsity from deployable sparse-kernel claims: parameter sparsity and FLOP accounting are useful compression proxies, but do not by themselves imply latency or throughput gains on a given backend \cite{gale2019state,blalock2020state,hoefler2021sparsity}.

\begin{table}[!htbp]
\centering
\caption{Closest ImageNet-1k unstructured and structured-sparsity context for the main rows. This compact table keeps the most comparable representation classes from the full comparison table: unstructured/fine-grained pruning, \(2{:}4\) or \(N{:}M\) semi-structured sparsity, and audited endpoint rows. The complete literature-context table is provided in Online Resource 2. Rows labeled ``ours'' are this paper's measurements; ``FT/train'' is the reported training or post-pruning fine-tuning budget.}
\label{tab:result_comparison}
\scriptsize
\setlength{\tabcolsep}{2pt}
\renewcommand{\arraystretch}{0.87}
\begin{tabularx}{\textwidth}{>{\raggedright\arraybackslash}p{2.35cm}
                  >{\raggedright\arraybackslash}p{1.9cm}
                  >{\raggedright\arraybackslash}p{2.3cm}
                  r r
                  >{\raggedright\arraybackslash}p{1.25cm}
                  X}
\toprule
Method & Architecture & Compression & Top-1 & \(\Delta\) & FT/train & Main comparison point \\
\midrule
Hybrid Mag--SER (ours) & ViT-B/16 & 50\% unstructured & \(83.37\) & \(-1.74\) & 1/cycle & MP-budgeted unstructured row with short-cycle FT. \\
CAST 2:4+ToMe (ours) & ViT-B/16 & \(2{:}4\)+ToMe & \(83.41\) & \(-1.70\) & 3 & Native A40 \(2{:}4\) endpoint \(1.388\times\); separate no-ToMe A100 endpoint \(2.705\times\). \\
SNOWS \cite{lucas2025snows} & ViT-B/16 & \(2{:}4\) QKV+Out+MLP & 76.57 & \(-3.85\) & 0 & Closest one-shot ViT \(2{:}4\) row; MiniImageNet-1k subset, no endpoint speed row. \\
MaskLLM (vision/4V) \cite{fang2024maskllm} & ViT-B/16 & \(2{:}4\) learned mask & 79.46 & \(+0.31\) & 20 mask ep. & Learned-mask ViT \(2{:}4\) context (MaskLLM vision/4V setting); weights frozen, no ViT endpoint speed reported. \\
SparseFormer \cite{gao2024sparseformer} & ViT-B/16 AugReg & latent-token reduction & 83.40 & \(-1.20\) & 20+5 ep. & Similar accuracy but token reduction, not weight sparsity; \(1.85\times\) throughput. \\
CAST 8:16 dense+perm (ours) & ViT-L/16 & 8:16 = 50\% & \(85.33\) & \(-0.51\) & 3 & Main wider-pattern accuracy/MAC row; no native 8:16 endpoint audit. \\
ToMe \cite{bolya2023tome} & ViT-L/16 MAE & token merging & 85.05 & \(-0.61\) & MAE FT & Closest token-merging accuracy/speed context; not weight sparsity. \\
UniPTS \cite{xie2024unipts} & ResNet-50 & 50/60/70\% unstructured & 75.76/75.37/74.73 & \(-0.36/-0.75/-1.39\) & 16k iters & Closest post-training unstructured ResNet row with limited calibration. \\
Hybrid Mag--SER (ours) & ResNet50 & 50/60/70\% unstructured & \(75.76/74.83/74.15\) & \(-0.37/-1.30/-1.98\) & 1/cycle & This paper's matched short-cycle unstructured ResNet sweep. \\
AC/DC \cite{peste2021acdc} & ResNet-50 & 50\% unstructured & 77.05 & \(+0.21\) & 100 train & Strong global sparse-training baseline; much longer training. \\
Mishra et al. \cite{mishra2021accelerating} & ResNet-50 & 2:4 FP16 & 76.20 & \(+0.10\) & repeated train & Original \(2{:}4\) sparse Tensor Core reference; up to \(2\times\) sparse math. \\
Pool--Yu perm. \cite{pool2021channel} & ResNet-50 & \(2{:}4\)+perm & 76.29 & \(+0.13\) & repeated train/FT & Closest channel-permutation semi-structured ResNet context (no inference-time overhead). \\
CAST-conv+perm (ours) & ResNet152d & 2:4 im2col & 81.33 & \(-1.53\) & 3 & Main CNN sparse-GEMM audit; \(1.617\times\) A40 im2col, not cuDNN Conv2d. \\
CAP \cite{kuznedelev2023cap} & ConvNeXt-L CLIP & 50/60/70\% unstructured & 87.5/87.1/86.8 & \(-0.3/-0.7/-1.0\) & 0 & Closest high-accuracy ConvNeXt unstructured one-shot context; no endpoint speed row. \\
Hybrid Mag--SER (ours) & ConvNeXtV2-B & 50/60/70\% unstructured & \(85.33/84.61/81.40\) & \(-1.39/-2.11/-5.32\) & 1/cycle & This paper's ConvNeXtV2 unstructured row with short-cycle FT. \\
CAST 12:16 (ours) & ConvNeXtV2-B & \(\sim\)25\% MAC-accounting & 86.35 & \(-0.37\) & 3 & Main ConvNeXtV2 wider-pattern accuracy/MAC row; no native 12:16 endpoint audit. \\
\bottomrule
\end{tabularx}
\end{table}

\subsection{Limitations and scope}
\label{subsec:limitations}

The main empirical scope qualifications are wider-pattern deployability, ResNet Conv2d endpoint support, artifact-bundle coverage, matched-ablation scope beyond ViT-B/16, and single-seed CAST row estimates; details are in Online Resource 2. The main theory limitation is the unestimated analytic \(L_s\) for modern transformer blocks; see Online Resource 1 for the local-Lipschitz discussion.

\FloatBarrier
\section{Abstract Additive Perturbation Extensions}
\label{sec:abstract_additive_extensions}

This section gives the abstract \(W=S+R\) additive analogue of the deterministic data-path framework in Section~\ref{Main_theory_gen}. Gaussian/RMT specializations and proofs are in Online Resource 1.

\subsection{Assumptions for the generalized perturbation framework}
\label{gen_ass_for_R}

Throughout the asymptotic statements in this section, the training set \(T\) is finite and fixed independently of \(N\), unless a result explicitly states summability conditions for a growing family.

\begin{assumption}[General architecture]
\label{as4}
Assume the DNN can be written as
\[
\phi=\rho\circ \psi_2\circ (R+S)\circ \psi_1,
\]
where \(\psi_1,\psi_2\) may depend on the remaining network parameters and \(\rho\) is softmax. For each \(s\in T\), let \(\mathcal U_s\) be any set containing the admissible interpolation segment \(\{(S+\vartheta R)\psi_1(s):\vartheta\in[0,1]\}\). We assume that \(\psi_2\) has a finite local Lipschitz constant with respect to \(\ell_\infty\) on \(\mathcal U_s\):
\[
L_{\psi_2}(s)
:=
\sup_{\substack{v,w\in\mathcal U_s\\ v\neq w}}
\frac{\|\psi_2(v)-\psi_2(w)\|_{\infty}}{\|v-w\|_{\infty}}
<\infty.
\]
\end{assumption}

\begin{assumption}[Perturbative random component]
\label{as5}
Assume that for every \(v\) measurable with respect to the conditioning variables \(\sigma(S,\psi_1,\psi_2)\), including deterministic \(v\),
\begin{equation}
\label{eq:general_perturb_assumption}
\PP\Big(\|Rv\|_{\infty}>d_1(N)J(v)\,\Big|\,S,\psi_1,\psi_2\Big)\le d_2(N),
\end{equation}
where \(d_1(N),d_2(N)\to0\) and \(J(v)\) depends only on \(v\). When a spectral interpretation is desired, we additionally assume that the singular-value ESD of \(R\) converges to a deterministic law with finite right edge \(b_+<\infty\).
\end{assumption}

For example, if \(R\in\mathbb R^{N\times N}\) has iid \(N(0,1/N)\) entries and \(v\) is fixed conditional on \(S,\psi_1,\psi_2\), then a Gaussian union bound gives Assumption~\ref{as5} with \(J(v)=\|v\|_2\), \(d_1(N)=2\sqrt{2\log(2N)/N}\), and \(d_2(N)=1/N\). Gaussian/RMT specializations and proofs are in Online Resource 1.

\begin{assumption}[Structured component for spectral conclusions]
\label{as6}
For any spectral conclusion, assume that there exists a sequence \(k_N=o(\min\{N,M\})\) such that
\[
\sum_{i>k_N}s_i(S)^2\to0.
\]
The convergence is in probability when \(S\) is random.
\end{assumption}

For a DNN satisfying Assumption~\ref{as4}, define
\begin{equation}
\label{from_DNN_st_2}
h_\phi(s):=J(\psi_1(s))\,L_{\psi_2}(s),
\qquad
a_\phi(N,s):=h_\phi(s)d_1(N).
\end{equation}

\subsection{Generalized perturbation and loss-reduction results}
\label{decrease_in_loss}

We consider the regularized loss
\begin{equation}
\label{loss_noise_det_2}
L(\alpha(t))
=
-\frac{1}{|T|}\sum_{s\in T}\log\big(\phi_{C(s)}(s,\alpha(t))\big)
+
\lambda_{\mathrm{reg}}
\sum_{i=1}^{L}\|W_i(t)\|_F^2.
\end{equation}

\begin{lem}
\label{lem:remove_R_general}
Let
\[
X=\psi_2\circ (R+S)\circ \psi_1(s),
\]
and let \(\alpha_W,\alpha_S\) denote the corresponding network parameters when the relevant weight matrix is \(W=R+S\) and \(S\), respectively. Under Assumptions~\ref{as4}--\ref{as5},
\[
\PP\Big(
\|X(s,\alpha_S)-X(s,\alpha_W)\|_\infty
\le d_1(N)h_\phi(s)
\Big)
\ge 1-d_2(N).
\]
\end{lem}

\begin{thm}
\label{theorem_reduction_of_loss_general}
Assume Assumptions~\ref{as4}--\ref{as5} and that
\[
\langle S,R\rangle_F\to0
\qquad\text{in probability as }N\to\infty.
\]
Assume further that for all \(s\in T\),
\[
a_\phi(N,s)\to0
\qquad\text{as }N\to\infty.
\]
Then removing \(R\) yields
\[
L(\alpha_S)=L(\alpha_W)-\lambda_{\mathrm{reg}}\|R\|_F^2+o_{\PP}(1)
\qquad(N\to\infty).
\]
\end{thm}

Secondary loss and finite-rank deformation corollaries, with proofs, are stated in Online Resource 1.

\begin{thm}[Multi-layer telescoping loss reduction]
\label{thm:multilayer_telescoping}
Let \(\mathcal P\) be a finite set of layers selected for pruning, with \(|\mathcal P|\) fixed independently of \(N\). For each \(\ell\in\mathcal P\), write
\[
W_\ell=S_\ell+R_\ell,
\]
and assume that, after the previous layers in \(\mathcal P\) have already been replaced by their structured parts, the single-layer hypotheses of Theorem~\ref{theorem_reduction_of_loss_general} continue to hold for layer \(\ell\) with perturbation scale \(a_{\phi,\ell}(N,s)\). If \(\alpha^{(0)}\) denotes the original network parameters and \(\alpha^{(|\mathcal P|)}\) the parameters after replacing every \(W_\ell\) by \(S_\ell\), then
\[
L(\alpha^{(|\mathcal P|)})
=
L(\alpha^{(0)})
-\lambda_{\mathrm{reg}}\sum_{\ell\in\mathcal P}\|R_\ell\|_F^2
+o_{\PP}(1).
\]
More generally, the set of pruned layers may depend on \(N\). Let \(\mathcal P_N\) be a sequence of layer sets, ordered in the pruning order. Assume that the corresponding one-layer logit events \(E_{\ell,N}\) satisfy
\[
\sum_{\ell\in\mathcal P_N}\PP(E_{\ell,N}^c)\to0,
\]
that their perturbation scales are summable,
\[
\sum_{\ell\in\mathcal P_N}\frac{1}{|T|}\sum_{s\in T}a_{\phi,\ell}(N,s)\to0,
\]
and that the accumulated cross term is negligible,
\[
\sum_{\ell\in\mathcal P_N}\langle S_\ell,R_\ell\rangle_F=o_{\PP}(1).
\]
Then the same telescoping conclusion holds with \(\mathcal P\) replaced by \(\mathcal P_N\).
\end{thm}

\begin{thm}
\label{main_result_reducing_noise_general}
Assume Assumptions~\ref{as4}--\ref{as5}. Assume further that for all \(s\in T\),
\[
a_\phi(N,s)\to0
\qquad\text{as }N\to\infty.
\]
Suppose:
\begin{enumerate}[leftmargin=2em]
\item for each \(N\), \(\alpha(t,N)\to \alpha^*(N)\) in probability as \(t\to\infty\);
\item the relevant limit weight matrix admits an admissible decomposition \(W^*(N)=S^*(N)+R^*(N)\) satisfying
\[
\langle S^*(N),R^*(N)\rangle_F\to0
\qquad\text{in probability as }N\to\infty.
\]
\end{enumerate}
\noindent For \(\vartheta\in[0,1]\), let \(\alpha_\vartheta^*(N)\) denote the parameter vector obtained by replacing \(W^*(N)\) with \(S^*(N)+\vartheta R^*(N)\). Assume further that there exists a sequence of nonnegative random variables \(\epsilon_N=o_{\PP}(1)\) such that the one-sided directional stationarity event
\[
\liminf_{h\downarrow0}
\frac{
L(\alpha_{1-h}^*(N))-L(\alpha^*(N))
}{h}
\ge -\epsilon_N
\]
has probability tending to one.
Define
\[
\bar a_{\phi,N}:=\frac{2}{|T|}\sum_{s\in T}a_\phi(N,s).
\]
Then
\[
\|R^*(N)\|_F^2
\le
\frac{\bar a_{\phi,N}+\epsilon_N}{2\lambda_{\mathrm{reg}}}
+o_{\PP}(1).
\]
In particular, if \(\bar a_{\phi,N}+\epsilon_N\to0\) in probability, then for every \(\epsilon>0\),
\[
\lim_{N\to\infty}\PP\big(\|R^*(N)\|_F\le \epsilon\big)=1.
\]

If, in addition, there exists an admissible path decomposition
\[
W(t,N)=S(t,N)+R(t,N)
\]
such that for each fixed \(N\),
\[
\|S(t,N)-S^*(N)\|_F+\|R(t,N)-R^*(N)\|_F\to0
\qquad\text{in probability as }t\to\infty,
\]
and if for every fixed \(N\) and every \(\epsilon>0\),
\[
\PP\big(\|R^*(N)\|_F=\epsilon\big)=0,
\]
then
\[
\lim_{N\to\infty}\lim_{t\to\infty}\PP\big(\|R(t,N)\|_F\le \epsilon\big)=1.
\]
\end{thm}

\subsection{Main Gaussian/RMT corollaries}
\label{subsec:main_gaussian_rmt_corollaries}

The following corollaries record the Gaussian/RMT specialization used in the MP interpretation. They are stated here because they are main mathematical consequences of the framework; their proofs and the detailed Gaussian assumptions are in Online Resource 1.

\begin{thm}[Gaussian stationarity collapse]
\label{thm:main_gaussian_stationarity_collapse}
Consider the square iid-Gaussian specialization of the additive framework for a trained limit layer
\[
W_2^*(N)=S_2^*(N)+R_2^*(N),
\qquad
R_2^*(N)_{ij}\sim N\!\left(0,\frac{g_*(N)}{N}\right),
\]
with the same local-path, cross-term, and one-sided stationarity hypotheses as Theorem~\ref{main_result_reducing_noise_general}. If the propagated perturbation scale and stationarity error vanish, then
\[
\|R_2^*(N)\|_F\to0
\qquad\text{in probability.}
\]
If an admissible training-path decomposition converges locally to this limit decomposition, then for every \(\epsilon>0\),
\[
\lim_{N\to\infty}\lim_{t\to\infty}
\PP\big(\|R_2(t,N)\|_F\le\epsilon\big)=1.
\]
\end{thm}
\proofloc{See Online Resource 1, proof of the Gaussian stationarity-collapse theorem.}

\begin{cor}[Variance-scale collapse at the limit point]
\label{cor:main_variance_scale_collapse}
Under the hypotheses of Theorem~\ref{thm:main_gaussian_stationarity_collapse}, the Gaussian variance scale satisfies
\[
g_*(N)N\to0.
\]
Consequently, in the square case the MP singular-value edge of the admissible limit perturbation satisfies
\[
\sigma_{+,*}^{\mathrm{bulk}}(N)=2\sqrt{g_*(N)}\to0.
\]
For a rectangular \(N\times M_N\) perturbation with entries \(N(0,g_*(N)/N)\), the corresponding Frobenius collapse condition is \(g_*(N)M_N\to0\), equivalent to \(g_*(N)N\to0\) when \(M_N/N\to c\in(0,\infty)\).
\end{cor}

\begin{cor}[Bulk collapse and spike stabilization at the limit point]
\label{cor:main_bulk_collapse_spike_stabilization}
Under the hypotheses of Theorem~\ref{thm:main_gaussian_stationarity_collapse}, assume the structured component is spectrally admissible, with singular values \(\sigma_i^*(N)\) and approximate rank scale \(k_N=o(N)\). If \(s_i(W_2^*(N))\) are the singular values of the full layer, then \(\|R_2^*(N)\|_{\op}\to0\) in probability, each fixed spike satisfies
\[
|s_i(W_2^*(N))-\sigma_i^*(N)|\to0
\qquad\text{in probability},
\]
the tail energy \(\sum_{i>k_N}s_i(W_2^*(N))^2\to0\) in probability, and the singular-value empirical distribution has no mass above any fixed \(\epsilon>0\) asymptotically. If \(\sigma_i^*(N)\to\bar\sigma_i\), then \(s_i(W_2^*(N))\to\bar\sigma_i\) for each fixed \(i\).
\end{cor}

\begin{cor}[Singular-subspace stabilization under an asymptotic gap condition]
\label{cor:main_subspace_stabilization}
Under the hypotheses of Theorem~\ref{thm:main_gaussian_stationarity_collapse}, fix \(k\ge1\) and let \(\delta_k(N)=\sigma_k^*(N)-\sigma_{k+1}^*(N)\). If \(\PP(\delta_k(N)>0)\to1\) and
\[
\frac{\|R_2^*(N)\|_{\op}}{\delta_k(N)}\to0
\qquad\text{in probability},
\]
then the top-\(k\) left and right singular-subspace projectors of \(W_2^*(N)\) converge in operator norm to those of \(S_2^*(N)\).
\end{cor}

\begin{cor}[Eventual Gaussian supercriticality of persistent spikes]
\label{cor:main_gaussian_BBP_persistent_spikes}
Under the hypotheses of Theorem~\ref{thm:main_gaussian_stationarity_collapse}, assume a square finite-rank Gaussian deformation regime and fixed persistent spikes
\[
\sigma_i^*(N)\to\bar\sigma_i>0,
\qquad 1\le i\le k.
\]
Then \(\PP(\sigma_i^*(N)>\sqrt{g_*(N)})\to1\), so these spikes are eventually supercritical above the Gaussian bulk edge \(2\sqrt{g_*(N)}\). With the usual spike-separation condition, the corresponding spectral projectors of \(W_2^*(N)\) converge to those of \(S_2^*(N)\), and the inverse-BBP estimator
\[
\widehat\sigma_i^{\mathrm{BBP}}(N)
:=
\frac12\left(s_i(W_2^*(N))+
\sqrt{s_i(W_2^*(N))^2-4g_*(N)}\right)
\]
is consistent in probability.
\end{cor}

\FloatBarrier
\section{Main Theoretical Results: Deterministic Pruning Certificates}
\label{Main_theory_gen}

This section states deterministic data-path certificates based on \(B_T(R)\). The mask-specific results concern fixed kept/removed supports and do not assume an additive Gaussian decomposition; random models only supply sufficient bounds for that budget. Proofs are collected in Online Resource 1.

\subsection{Deterministic data-path certificates for mask pruning}
\label{subsec:deterministic_mask_certificate}

Fix one target layer \(A\in\R^{n\times m}\) inside a network and write the logits as
\[
z_A(s):=\psi_2(A\psi_1(s))\in\R^K,
\qquad s\in T,
\]
where \(\psi_1\) and \(\psi_2\) denote the pre-layer and post-layer computations with all other parameters frozen. For a decomposition
\[
A=S+R,
\qquad
A_\theta:=S+\theta R,\qquad 0\le\theta\le1,
\]
let \(L_s\) be any finite deterministic upper bound on the local \(\ell_\infty\)-Lipschitz constant of \(\psi_2\) on the path
\[
\{A_\theta\psi_1(s):0\le\theta\le1\}.
\]
Define the data-path budget of the removed component by
\begin{equation}
\label{eq:data_path_budget}
B_T(R):=
\frac{2}{|T|}\sum_{s\in T}L_s\|R\psi_1(s)\|_\infty.
\end{equation}
If \(L_s\) is itself estimated from the realized network, the statements below are deterministic conditional on that realized upper bound.

\begin{lem}[Deterministic cross-entropy path bound]
\label{lem:deterministic_ce_path_bound}
For every decomposition \(A=S+R\) and every \(\theta\in[0,1]\),
\[
\left|
\mathcal L_{\mathrm{CE}}(A_\theta)-\mathcal L_{\mathrm{CE}}(A_1)
\right|
\le
(1-\theta)B_T(R),
\]
where
\[
\mathcal L_{\mathrm{CE}}(A)
:=
\frac1{|T|}\sum_{s\in T}
\left[
-z_A(s)_{C(s)}
+\log\!\left(\sum_{j=1}^K e^{z_A(s)_j}\right)
\right].
\]
\end{lem}

\begin{cor}[Deterministic margin stability]
\label{cor:deterministic_margin_stability}
Let
\[
\eta_s(\theta):=(1-\theta)L_s\|R\psi_1(s)\|_\infty.
\]
If
\[
p_A(s)\in\arg\max_j z_A(s)_j,
\qquad
\Delta_A^{\mathrm{pred}}(s):=
z_A(s)_{p_A(s)}-\max_{j\ne p_A(s)}z_A(s)_j,
\]
then the predicted label at \(s\) can change between \(A_1\) and \(A_\theta\) only if
\[
\Delta_{A_1}^{\mathrm{pred}}(s)\le 2\eta_s(\theta).
\]
For the true-label margin
\[
\Delta_A^{\mathrm{true}}(s):=
z_A(s)_{C(s)}-\max_{j\ne C(s)}z_A(s)_j,
\]
one has
\[
\big|\acc_{A_\theta}(T)-\acc_{A_1}(T)\big|
\le
\frac1{|T|}
\#\left\{s\in T:
\left|\Delta_{A_1}^{\mathrm{true}}(s)\right|
\le 2\eta_s(\theta)
\right\}.
\]
Here
\[
\acc_A(T):=\frac1{|T|}\#\{s\in T:\Delta_A^{\mathrm{true}}(s)>0\}.
\]
\end{cor}

\begin{thm}[Deterministic additive Frobenius certificate]
\label{thm:deterministic_additive_frobenius_certificate}
For any decomposition \(A=S+R\), not necessarily a mask decomposition, define the one-layer Frobenius-regularized objective
\[
\mathcal J_2(A):=
\mathcal L_{\mathrm{CE}}(A)+\lambda_2\|A\|_F^2+\mathcal C,
\qquad \lambda_2\ge0,
\]
where \(\mathcal C\) contains all terms independent of \(A\). Then for every \(\theta\in[0,1]\),
\[
\mathcal J_2(A_\theta)-\mathcal J_2(A_1)
\le
(1-\theta)B_T(R)
-
\lambda_2(1-\theta^2)\|R\|_F^2
-
2\lambda_2(1-\theta)\langle S,R\rangle_F.
\]
Consequently, if \(|\langle S,R\rangle_F|\le\eta\), then
\[
\mathcal J_2(A_\theta)-\mathcal J_2(A_1)
\le
(1-\theta)B_T(R)
-
\lambda_2(1-\theta^2)\|R\|_F^2
+
2\lambda_2(1-\theta)\eta.
\]
In particular, full removal is certified to decrease the Frobenius-regularized objective whenever
\[
B_T(R)+2\lambda_2|\langle S,R\rangle_F|
<
\lambda_2\|R\|_F^2.
\]
\end{thm}

\begin{thm}[Deterministic elastic-net mask pruning certificate]
\label{thm:deterministic_mask_certificate}
Assume \(A=S+R\) is a mask decomposition, meaning
\[
S\odot R=0.
\]
For constants \(\lambda_2,\lambda_1\ge0\), define the one-layer elastic-net objective by
\[
\mathcal J(A):=
\mathcal L_{\mathrm{CE}}(A)
+\lambda_2\|A\|_F^2
+\lambda_1\|A\|_1
+\mathcal C,
\]
where \(\mathcal C\) contains all terms independent of \(A\). Then for every \(\theta\in[0,1]\),
\[
\mathcal J(A_\theta)-\mathcal J(A_1)
\le
(1-\theta)B_T(R)
-\lambda_2(1-\theta^2)\|R\|_F^2
-\lambda_1(1-\theta)\|R\|_1.
\]
For full pruning,
\[
\mathcal J(S)
\le
\mathcal J(A)
-\lambda_2\|R\|_F^2
-\lambda_1\|R\|_1
+B_T(R).
\]
Consequently, the mask is certified to decrease the elastic-net objective whenever
\[
B_T(R)
<
\lambda_2\|R\|_F^2+\lambda_1\|R\|_1.
\]
\end{thm}

\begin{thm}[Deterministic prune--restore certificate]
\label{thm:deterministic_prune_restore_certificate}
Let the original layer decompose as
\[
A=S+Q+P,
\qquad
S\odot Q=S\odot P=Q\odot P=0.
\]
Interpret \(P\) as the entries finally removed and \(Q\) as entries restored, or kept, from an over-pruned reservoir. Define
\[
A_\theta:=S+Q+\theta P,\qquad 0\le\theta\le1.
\]
For the elastic-net objective \(\mathcal J\) of Theorem~\ref{thm:deterministic_mask_certificate}, one has
\[
\mathcal J(A_\theta)-\mathcal J(A)
\le
(1-\theta)B_T(P)
-\lambda_2(1-\theta^2)\|P\|_F^2
-\lambda_1(1-\theta)\|P\|_1.
\]
For the final prune--restore layer \(S+Q\),
\[
\mathcal J(S+Q)
\le
\mathcal J(A)
-\lambda_2\|P\|_F^2
-\lambda_1\|P\|_1
+B_T(P).
\]
Consequently, the final SER-style mask is certified to decrease the elastic-net objective whenever
\[
B_T(P)<\lambda_2\|P\|_F^2+\lambda_1\|P\|_1.
\]
\end{thm}

\begin{cor}[When restoration improves the certificate upper bound]
\label{cor:restoration_improves_certificate}
Under the hypotheses of Theorem~\ref{thm:deterministic_prune_restore_certificate}, set \(R:=Q+P\). Comparing the all-pruned certificate for \(A=S+R\) with the restore-\(Q\) certificate for \(A=(S+Q)+P\), the restored upper bound is strictly smaller iff
\[
B_T(R)-B_T(P)
>
\lambda_2\|Q\|_F^2+\lambda_1\|Q\|_1.
\]
\end{cor}

Coordinatewise weighted variants are stated in Online Resource 1.

\begin{thm}[One-sided stationarity controls removable masks]
\label{thm:deterministic_mask_stationarity}
Under the hypotheses of Theorem~\ref{thm:deterministic_mask_certificate}, assume the one-sided directional stationarity condition holds for some \(\varepsilon\ge0\):
\[
\liminf_{h\downarrow0}
\frac{\mathcal J(A_{1-h})-\mathcal J(A_1)}{h}
\ge
-\varepsilon.
\]
Then
\[
2\lambda_2\|R\|_F^2+\lambda_1\|R\|_1
\le
B_T(R)+\varepsilon.
\]
More generally, if
\[
\mathcal J(A_\theta)\ge \mathcal J(A_1)-\varepsilon(1-\theta)
\qquad\forall\theta\in[0,1],
\]
then for every \(\theta<1\),
\[
\lambda_2(1+\theta)\|R\|_F^2+\lambda_1\|R\|_1
\le
B_T(R)+\varepsilon.
\]
\end{thm}

\begin{thm}[Deterministic multi-layer mask certificate]
\label{thm:deterministic_multilayer_mask_certificate}
Let \(\mathcal J(\alpha)\) denote the full cross-entropy plus layerwise elastic-net objective. Consider an ordered finite sequence of mask-pruning operations on layers \(j=1,\dots,M\). At step \(j\), write the current layer after steps \(1,\dots,j-1\) as
\[
A_j=S_j+R_j,
\qquad
S_j\odot R_j=0,
\]
and let \(B_j(R_j)\) be its data-path budget in that intermediate network. If \(\alpha^{(0)}\) and \(\alpha^{(M)}\) denote the parameters before and after the \(M\) mask removals, then
\[
\mathcal J(\alpha^{(M)})
\le
\mathcal J(\alpha^{(0)})
-
\sum_{j=1}^M\lambda_{2,j}\|R_j\|_F^2
-
\sum_{j=1}^M\lambda_{1,j}\|R_j\|_1
+
\sum_{j=1}^M B_j(R_j),
\]
where \(\lambda_{2,j}\) and \(\lambda_{1,j}\) are the elastic-net weights applied to layer \(j\).
\end{thm}

The corresponding multi-layer margin-stability corollary is stated in Online Resource 1.

\begin{thm}[Multi-layer stationarity controls jointly removable masks]
\label{thm:deterministic_multilayer_mask_stationarity}
Use the notation of Theorem~\ref{thm:deterministic_multilayer_mask_certificate}. For \(h\in[0,1]\), let \(\alpha_h\) be obtained by applying the same ordered pruning operations with each current layer \(A_j=S_j+R_j\) replaced by \(S_j+(1-h)R_j\). Assume that each \(B_j(R_j)\) remains valid along this fractional path, uniformly for \(h\in[0,1]\), so the step-\(j\) cross-entropy increase is at most \(hB_j(R_j)\). If the one-sided directional stationarity condition
\[
\liminf_{h\downarrow0}
\frac{\mathcal J(\alpha_h)-\mathcal J(\alpha^{(0)})}{h}
\ge
-\varepsilon
\]
holds for some \(\varepsilon\ge0\), then
\[
2\sum_{j=1}^M\lambda_{2,j}\|R_j\|_F^2
+
\sum_{j=1}^M\lambda_{1,j}\|R_j\|_1
\le
\sum_{j=1}^M B_j(R_j)+\varepsilon.
\]
More generally, if
\[
\mathcal J(\alpha_h)\ge\mathcal J(\alpha^{(0)})-\varepsilon h
\qquad\forall h\in[0,1],
\]
then for every \(h\in(0,1]\),
\[
\sum_{j=1}^M\lambda_{2,j}(2-h)\|R_j\|_F^2
+
\sum_{j=1}^M\lambda_{1,j}\|R_j\|_1
\le
\sum_{j=1}^M B_j(R_j)+\varepsilon.
\]
\end{thm}

\begin{cor}[Gaussian sufficient condition for the data-path budget]
\label{cor:gaussian_data_path_budget}
Let \(R\in\R^{n\times m}\) have independent entries \(R_{ij}\sim N(0,g/n)\), independent of the collection \(v_s:=\psi_1(s)\), \(s\in T\). Assume \(L_s\) are nonnegative upper bounds for the post-layer Lipschitz constants, deterministic or measurable with respect to variables independent of \(R\). Then for every finite \(T\) and every \(\delta\in(0,1)\), with probability at least \(1-\delta\),
\[
B_T(R)
\le
\frac2{|T|}
\sum_{s\in T}
L_s\|v_s\|_2
\sqrt{
\frac{2g}{n}\log\frac{2n|T|}{\delta}
}.
\]
\end{cor}

\begin{cor}[MP-edge form of the Gaussian data-path budget]
\label{cor:mp_edge_data_path_budget}
Under the hypotheses of Corollary~\ref{cor:gaussian_data_path_budget}, define the nominal rectangular MP singular-value edge of \(R\in\R^{n\times m}\) by
\[
\sigma_{+,R}:=\sqrt g\left(1+\sqrt{\frac{m}{n}}\right).
\]
Then, with probability at least \(1-\delta\),
\[
B_T(R)
\le
\frac{2\sigma_{+,R}}{1+\sqrt{m/n}}\,
\sqrt{\frac{2}{n}\log\frac{2n|T|}{\delta}}\,
\frac1{|T|}\sum_{s\in T}L_s\|v_s\|_2.
\]
\end{cor}

Sub-Gaussian and simultaneous multi-layer budget corollaries are stated in Online Resource 1.

\subsection*{Acknowledgments}
LB, HO, and YS acknowledge support from NASA via the AIST program (Kernel Flows: Emulating Complex Models for Massive Data Sets). This work started during LB's sabbatical stay at Caltech hosted by H.\ Owhadi; LB and YS are grateful for the hospitality during their NASA-supported visit.

\section*{Statements and Declarations}

\subsection*{Funding}
The work of LB was partially supported by NSF grant DMS-2005262 and NSF grant IMPRESS-U 2401227. TB and HO acknowledge support from the Air Force Office of Scientific Research under MURI award number FA9550-20-1-0358 (Machine Learning and Physics-Based Modeling and Simulation) and FOA-AFRL-AFOSR-2023-0004 (Mathematics of Digital Twins), and from the Department of Energy under award number DE-SC0023163 (SEA-CROGS: Scalable, Efficient, and Accelerated Causal Reasoning Operators, Graphs and Spikes for Earth and Embedded Systems). HO and TB also acknowledge support from the DoD Vannevar Bush Faculty Fellowship Program under award number ONR-N000142512035. The work of YS was partially supported by the European Research Council under the European Union's Horizon 2022 research and innovation program (grant agreement No.~101041711), the Simons Foundation, the Israel Science Foundation (grant number 2258/19), and the Israel Science Foundation (ISF Grant~4101/25).

\subsection*{Competing interests}
The authors declare that they have no competing interests relevant to the content of this article.

\subsection*{Author contributions}
All authors contributed to the mathematical development, interpretation of results, manuscript revision, and approval of the submitted version.

\subsection*{Data, code, and supplementary information availability}
Code, plotting scripts, pruning recipes, manuscript artifacts, and mirror copies of Online Resource 1 and Online Resource 2 are available at \repoURL. The checkpoints, validation ledgers, sparse-backend timing evidence, and per-run files supporting the main ImageNet-1k tables are available at
\url{https://drive.google.com/drive/folders/1mm990SHAHlYdISHxirvMRdVQEAjpIxDd}. ImageNet-1k images are not redistributed; ImageNet is identified by \cite{5206848}, and users must obtain the dataset under the official download terms \cite{imagenet_download_terms}. Online Resource 2 identifies the source checkpoints, recipes, evidence files, timing audits, and complete numerical ledgers for the reported rows.

\subsection*{Declaration of generative AI and AI-assisted technologies in the manuscript preparation process}
During the preparation of this work the author(s) used generative AI tools to accelerate drafting and revision. All mathematical claims, proofs, numerical interpretations, and bibliographic information were subsequently reviewed and edited by the author(s), who take full responsibility for the content of the manuscript.

\bibliographystyle{plainnat}
\bibliography{rmt_vit_pruning}

\end{document}